\documentclass[10pt]{article}
\usepackage{amsmath}
\usepackage{amsfonts}
\usepackage{amssymb}
\usepackage{mathrsfs}

\usepackage{fancyhdr}
\usepackage{blindtext}

\usepackage{MnSymbol}

\usepackage[cal=boondox,scr=boondoxo]{mathalfa}

\usepackage{float}
\usepackage{subfig}
\usepackage{fancybox,graphicx}
\usepackage{subfig}
\usepackage{caption}
\usepackage{color}
\usepackage{authblk}

\usepackage[colorlinks]{hyperref}
\usepackage{hyperref}

\newcommand{\doi}[1]{{doi:~\href{https://doi.org/#1}{\nolinkurl{#1}}}\rmFullStop}

\newcommand*{\rmFullStop}{\rmifnextchar{.}{}{}}

\makeatletter
\newcommand{\rmifnextchar}[3]{%
  \begingroup
  \ltx@LocToksA{\endgroup#2}%
  \ltx@LocToksB{\endgroup#3}%
  \ltx@ifnextchar{#1}{%
    \def\next{\the\ltx@LocToksA}%
    \afterassignment\next
    \let\scratch= %
  }{%
    \the\ltx@LocToksB
  }%
}
\makeatother %doi command

\usepackage{accents}
\usepackage[titletoc,title]{appendix}
\usepackage{cite}

\usepackage{mathtools}

\usepackage[top=2in, bottom=1.5in, left=1in, right=1in]{geometry}

\usepackage{stackengine}

\title{Conditional Similarity Triplets Enable Covariate-Informed Representations of Single-Cell Data} 
 
\author[1,*]{Chi-Jane Chen}
\author[1]{Haidong Yi}
\author[1,2,3,*]{Natalie Stanley}

\affil[1]{Department of Computer Science, The University of North Carolina at Chapel Hill, Chapel Hill, NC 27599, USA}
\affil[2]{Computational Medicine Program, The University of North Carolina at Chapel Hill, Chapel Hill, NC 27599, USA}
\affil[3]{Department of Genetics, The University of North Carolina at Chapel Hill, Chapel Hill, NC 27599, USA}
\affil[*]{Corresponding authors: chijane@cs.unc.edu,  natalies@cs.unc.edu}
\begin{document}

\maketitle
\thispagestyle{fancy}
\begin{abstract}
\noindent Single-cell technologies enable comprehensive profiling of diverse immune cell-types through the measurement of multiple genes or proteins per individual cell. In order to translate immune signatures assayed from blood or tissue into powerful diagnostics, machine learning approaches are often employed to compute immunological summaries or per-sample \emph{featurizations}, which can be used as inputs to models for outcomes of interest. Current supervised learning approaches for computing per-sample representations are trained only to accurately predict a single outcome and do not take into account relevant additional clinical features or covariates that are likely to also be measured for each sample. Here, we introduce a novel approach for incorporating measured covariates in optimizing model parameters to ultimately specify per-sample encodings that accurately affect both immune signatures and additional clinical information. Our introduced method CytoCoSet is a set-based encoding method for learning per-sample featurizations, which formulates a loss function with an additional triplet term penalizing samples with similar covariates from having disparate embedding results in per-sample representations. Overall, incorporating clinical covariates enables the learning of encodings for each individual sample that ultimately improve prediction of clinical outcome.
\end{abstract}
\section{Introduction}
%intro to single cell + application
High-throughput single-cell technologies, such as flow and mass cytometry and single-cell RNA sequencing are prominent experimental technologies for unraveling key genetic programs and cell-types driving particular disease states or biological processes \cite{bendall2012deep,brodin2019call,jagadeesh2022identifying}. In doing so, it is common to measure the expression of multiple molecular features in each individual cell \cite{regev2017human} to ultimately phenotype cells \cite{perplexed} and to characterize their functional responses in disease states \cite{chen2022graph,cui2024dictionary}. The era of high throughput immune profiling has shown that changes in the immune system can be prominent drivers of certain disease states \cite{davis2017systems}. For example, certain characteristic and chronological changes in frequencies of immune cell-types have been observed during the course of a healthy pregnancy, reflecting key dynamic adaptations in the maternal immune system over time {aghaeepour2017immune}. In this pregnancy example, as well as in numerous other applications \cite{peterson2021single,olin2018stereotypic,vanderbeke2021monocyte,lo2022molecular}, such changes in frequencies (e.g. relative abundances of immune cell-types) can be used as input features to build machine learning models of diseases or dynamic processes. Similarly, clinically-associated functional changes, such as the activation of particular signaling pathways, across cells are also easily interrogated through single-cell immune profiling techniques. For example, a recent study uncovered differing levels of activation of particular signaling pathways in certain immune cell types between patients who did and did not receive a glucocorticoid treatment after surgery \cite{ganio2020preferential}. 

%problem: predicting outcomes from sc
%related worked cellcnn,cytoset,cytoDX,frequency features engineering(kmean), RFF
To accommodate the information-rich datasets generated by single-cell technologies, machine learning methods are crucial for linking complex patterns in cell-type abundances and their associated gene expression programs to clinical or experimental phenotypes. In formulating a concrete clinical prediction task, there are numerous approaches for specifying quality per-sample feature encodings or \emph{featurizations} to adequately summarize cellular correlates for the outcome of interest. Moreover, such featurizations must be information-rich enough to capture subtle patterns in between sample heterogeneity, and align with the assumption that feature encodings are, on average, more similar between samples with the same clinical outcome (label) than those with different labels. Broadly, featurization approaches are implemented by using either 1) biologically-meaningful per-sample feature engineering techniques or by 2) mathematically abstract, yet detailed feature encodings, often inferred through a deep learning approach.
%next sction of this paragraph is talking about non deep learning based approaches (e.g., gating based + feature engineering and CytoDX)

As initial strategies to translate cellular heterogeneity patterns into immune features, Refs. \cite{vopo,bruggner2014automated} introduce \emph{gating-based approaches} for automated featurization and classification from single-cell profiled samples. These approaches specify cluster-level frequency features in each sample by clustering cells into populations and counting the number of cells in each cluster. Although biologically interpretable, cluster-based feature engineering strategies are sensitive to the number of clusters chosen and are subject to variability across clustering runs. 
% Two examples of biologically-interpretable feature representations include frequency feature engineering\cite{stanley2020vopo} and CytoDX \cite{hu2019robust}.  
%The next part of this paragraph is talking about learning based feature engineering approaches, or those that are not biologically interpretable (cellCNN, RFF)
The other class of techniques are referred to as feature-learning approaches, whereby a mathematical abstraction is learned for each sample, but lacks biological interpretability. These methods include CellCNN \cite{arvaniti2017sensitive}, Random Fourier-based featurization \cite{rahimi2007random}, CytoDX \cite{hu2019robust}, and CytoSet \cite{yi2021cytoset}. CytoSet \cite{yi2021cytoset} and CellCNN \cite{arvaniti2017sensitive} are deep-learning-based approach for encoding complex patterns in multiple CyTOF samples, which incorporate the outcome of interest \cite{yi2021cytoset} in learning model parameters. Specifically, CytoSet is most similar to our work and learns featurizations or embeddings for each sample in a single-cell dataset by encoding each sample's molecular profiles into fixed-length vectors using permutation-invariant neural networks. Broadly, CytoSet ensures that embeddings capture sample-level patterns in cell-type composition, and patterns per-cell feature expression. Although CytoSet successfully predicts clinical outcomes based on immune signatures, some datasets have complex interplays between particular immune cell-types, clinical covariates and the outcome of interest, making prediction more challenging. Finally, additional feature learning approaches that do not use a deep learning approach include CytoDX \cite{hu2019robust} and CKME \cite{shan2022transparent}, which use regression and kernel based feature learning, respectively. CytoDx is another method for predicting clinical outcomes from single-cell data in a gating-free manner\cite{hu2019robust}, and ultimately averages per-cell predictions across cells in the sample to get a sample-level prediction. 

%limitation of method % summary of my method + contribution
As alluded to in the discussion of CytoSet, these various featurization techniques can translate complex cellular landscape patterns into useful feature encodings, but they are agnostic to clinical covariates that are also available for each sample. Moreover, it is possible that certain clinical covariates may exhibit correlations with immune signatures and ultimately confound the prediction of the clinical outcome. To address these limitations, we propose CytoCoSet, a deep-learning-based model that utilizes patient covariates that are distinct from the outcome to be predicted in order to learn accurate sample-level feature encodings.  This enhanced, covariate-informed model enables sample-based feature representation learning or featurization, while accommodating for diverse covariates to ultimately generate clinically-holistic summaries of the immune system and the patient's background health. Further, our novel approach extends CytoSet \cite{yi2021cytoset}, which treats a multi-sample single-cell dataset as a dataset of sets, enabling set-based encoding using a permutation invariant network architecture \cite{zaheer2017deep}. As a motivating example of a clinical setting which should leverage patient covariates in conjunction with the immune system in clinical settings,  Peterson et al. showed that immune profiles from chord blood complemented with additional covariates, such as preeclampsia status and miscarriage history can be integrated to accurately predict deleterious pregnancy outcomes\cite{peterson2021single}.  Moreover, there are currently no algorithmic frameworks for integrating such features in an automated way with immune signatures. To enhance per-sample immunological summaries that take into account sample-level covariates (e.g. extra clinical features), we formulate an extension to the original CytoSet loss function, which includes an extra penalty term to enforce embeddings to be similar between samples with similar covariates. Overall, the integration of diverse clinically relevant summaries has the potential to enhance the accuracy and diagnostic power of immune-based diagnostics assayed through single-cell profiling. 
\begin{figure*}[!t]%
\includegraphics[width=1\textwidth]{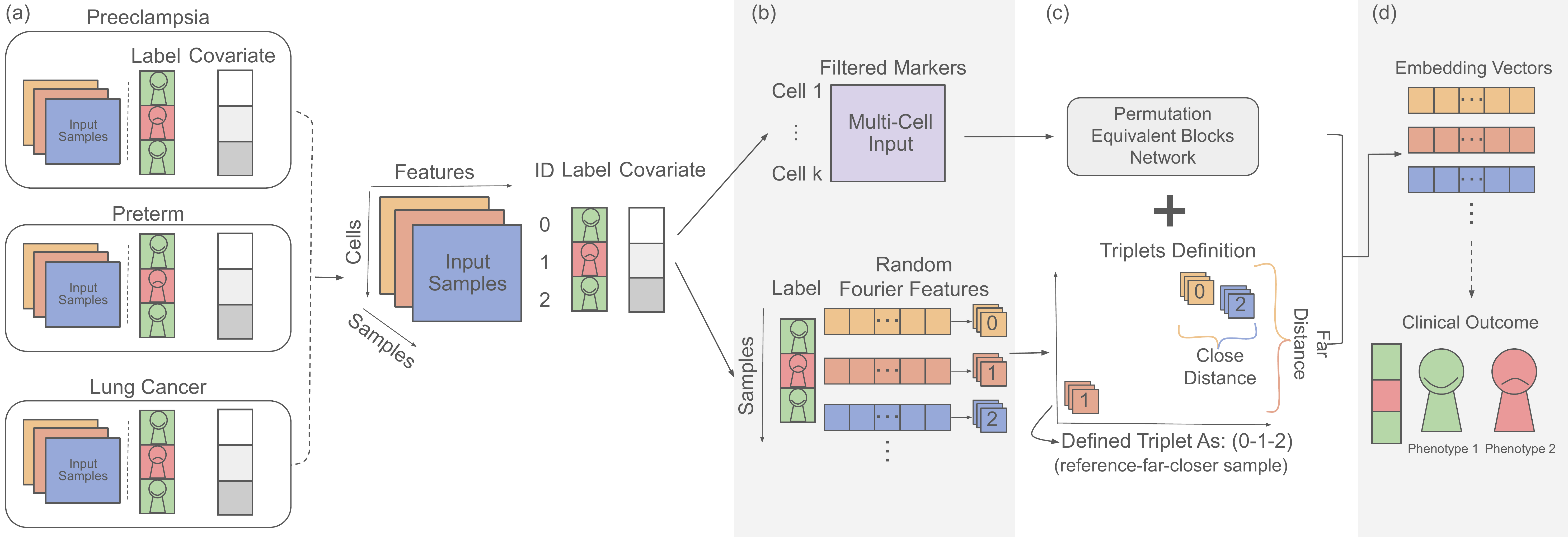}
\centering
\caption{\textbf{Overview.} Schematic overview of CytoCoSet. ({\bf a}) Given a multi-sample single-cell dataset additional covariates measured in each sample, ({\bf b})the CytoCoSet algorithm defines a set of triplets based on Random Fourier Features (RFFs) to constrain the process of learning per-sample embedding vectors. A triplet is a combination of three samples, such that two samples have similar covariates and should have similar embeddings, and the third sample is distinct in terms of covariates and should therefore have a more divergent embedding. ({\bf c}) The loss function specified to optimize the embeddings is comprised of a binary cross entropy term to enforce prediction accuracy, and a triplet term, which encodes covariate-based similarity constraints. ({\bf d}) Embedding vectors learned by the model can be used to train machine learning models of clinical outcome.}
\label{fig1}
\end{figure*}

\section{Methods}
CytoCoSet is a deep-learning based feature learning approach for encoding complex immunological and clinical feature patterns into sample-level featurizations or encodings that are predictive of clinical outcome. CytoCoSet innovatively augments the traditional supervised per-sample representation learning problem to also take into account the rich set of patient-specific covariates that are likely driving between-sample variation and outcome. Briefly, we formulated the loss as an extension to the set-based encoding method, CytoSet \cite{yi2021cytoset}, which optimizes parameters to ultimately compute per-sample representations that are predictive of clinical outcome. CytoCoSet extends the per-sample feature learning approach by augmenting the original loss function with a triplet term, which enforces samples with similar covariates to have more similar learned embeddings, similar to that described in Ref.\cite{veit2017conditional}. In applying this approach, we anticipate that the learned embeddings can identify novel groups of samples with highly similar immune features and covariates within each clinical outcome group. In the subsequent section, we describe the key components of the algorithm, including the loss function, optimal triplet selection based on Random Fourier Features (RFFs) and the model architecture of the deep network. \\

\begin{figure}[!t]%
\includegraphics[width=1\textwidth]{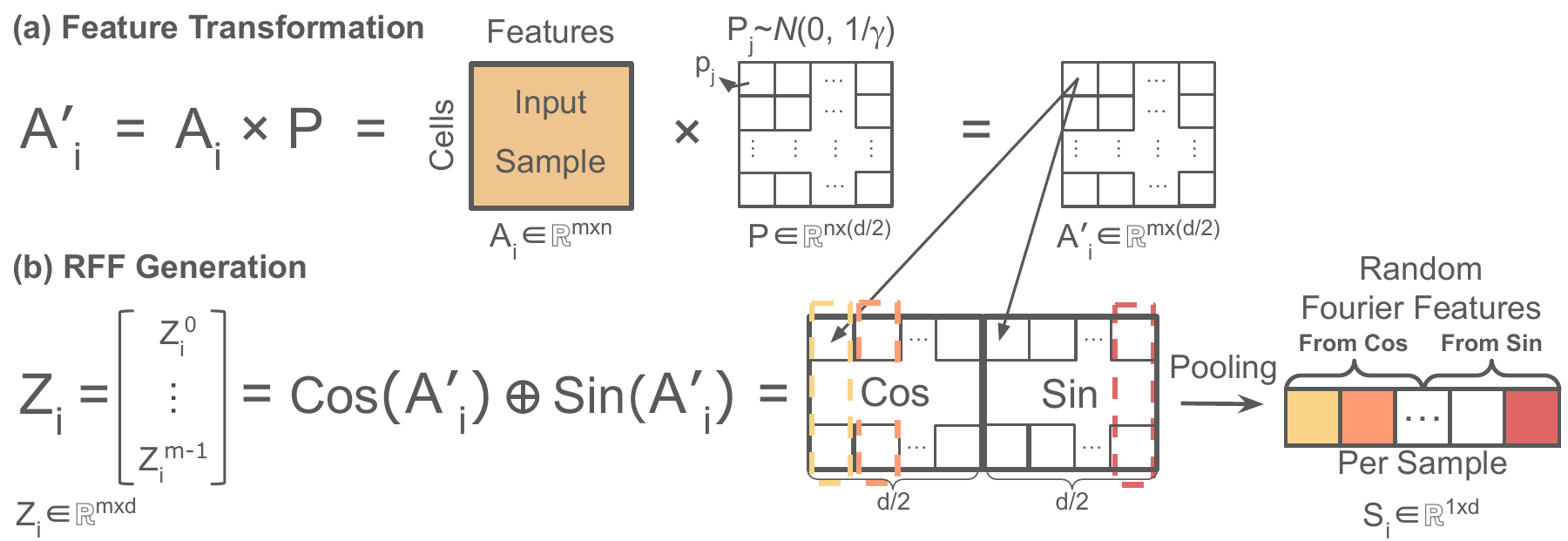}
\centering
\caption{\textbf{Overview of how Random Fourier Features are used to select triplets.} An illustration of how Random Fourier Features (RFFs) are used to summarize the overall immune landscape for each sample. (a) The columns of a cell $\times$ feature input matrix ${A}_i$ are transformed with $\frac{d}{2}$ Gaussian random variables ($P$) to produce a new matrix, ${A'}_i$. (b) Per-cell Random Fourier Features (RFFs) are constructed by concatenating sine and cosine transformed features for each cell (We use $\oplus$ as the notation for concatenation). Sine and cosine transformed values of ${A'}_i$ are used to form a matrix $Z_i$ across all cells. Finally, a Random Fourier Feature vector $S_{i}$ is constructed pooling (median or max) the feature values in each dimension across all cells. }\label{fig2}
\end{figure}

\noindent \textbf{Computing Random Fourier Feature Representations for Each Sample.} We specify triplets by choosing combinations of three samples, such that two samples of the pair are similar based on the covariate and two samples of the pair are distinct based on a measured covariate. Further, we aim to specify triplets that are maximally information-rich and produce a strong effect on the loss. Specifically, we used an unsupervised per-sample embedding approach \cite{shan2022transparent} to summarize intricate cellular patterns in an abstract vector representation based on Random Fourier Features (RFFs) \cite{rahimi2007random}. Briefly, Random Fourier Features infer mathematically abstract $d$-dimensional feature encodings per cell, which can be averaged across all cells in a sample to create a per-sample encoding. RFFs can be thought of as primitive, unsupervised encodings of complex non-linear effects in data that are often captured through deep learning approaches. In practice, we compute the Random Fourier Feature representation for $i$th sample (shown in Fig. \ref{fig2}), ${Z}_i \in \mathbb {R}^{m \times d}$, by approximating the feature map of the Gaussian kernel, as described in Refs. \cite{shan2022transparent,rahimi2007random}. First, we consider the $i$th sample's $m\times n$ single-cell data matrix $A_i$, which encodes the expressions of $n$ proteins in $m$ cells. Next, we generate a matrix, $P \in \mathbb{R}^{n \times d/2}$ Gaussian random variables, such that the $j$-th entry of the matrix, $P_{j} \sim \mathcal{N}(0,\frac{1}{\gamma})$.
%transform columns of $A_i$ based on $d/2$-dimensional randomly generated normal random variables. So, for a given column $j$ of ${A}_i$, we generate a transformation vector, $P_{j}$, of $n$ normally distributed random variables, such that each entry of $P_{j} \sim \mathcal{N}(0,\frac{1}{\gamma})$. We then concatenate the $P_{j}s$ row-wise to form an overall transformation matrix, $P \in \mathbb{R}^{n \times d/2}$ with $P=[P_{0} \mid P_{1} \mid \cdots \mid P_{n-1}]$. 
After generating the matrix $P$, we transform the input matrix $A_i$ into ${A'}_i$ as follows:  

\begin{equation}
\label{eq:1}
{A'}_i = A_i \times P.
\end{equation}

Next, we transform each row of ${A'}_i$ into a vector, such that the $q$th row of ${A'}_{i}$ yields a vector, $Z_{i}$, which evaluates the sine and cosine of each element. That is, for the $q$-th row of ${A'}_{i}$, we compute the vector $Z_{i}^{q}$ based on row $q$ and column $p$ of $A'_{i[q,p]}$ as, 

\begin{equation}
Z_{i}^{q}=\sqrt{\frac{2}{K}}[\cos({A'}_{i}{_{[q,0]}}),..., \cos({A'}_{i[q,\frac{d}{2}-1]}),\sin({A'}_{i[q,0]}),..., \sin({A'}_{i[q,\frac{d}{2}-1]}).
\end{equation}

Here, $K$ is the number of random features ($\frac{d}{2}$) used so, $\sqrt{\frac{2}{K}} $ is negligible. Moreover, we concatenate the individual $Z_{i}^{q}s$ by row to form a matrix, $Z_{i}$ of transformed representations across cells as,

\begin{equation}
\label{eq:stack}
Z_{i} = \begin{bmatrix}
Z_{i}^{0}\\
...\\
Z_{i}^{m-1}
\end{bmatrix}.
\end{equation}

Finally, to compress the matrix $Z_{i}$ (cell embedding per sample $i$) into a vector $S_i \in \mathbb {R}^{1 \times d}$, we apply a pooling operation (either median or max) each column of $Z_{i}$ as, 

\begin{equation}
\label{eq:pool}
S_i = \begin{cases} 
      \text{Pool}_{\text{median}}(Z_{i}) \\
      \text{Pool}_{\text{max}}(Z_{i}). 
   \end{cases}
\end{equation}
In practice, we found median pooling to be the most robust in our experiments. Ultimately, we use the per-sample representations, $S_i$, to compare the aggregate immune landscape between pairs of samples. \\

\noindent \textbf{Selecting Triplets Based on Random Fourier Features.} After defining encodings for each sample with Random Fourier Features (RFFs), we use these representations to specify triplets to improve model training. We define $S_{i}$ as the vector of Random Fourier Features (RFFs) for sample $i$. We specify each triplet based on three samples, which namely have indices $i$, $j$, and $k$, with associated RFFs $S_i$, $S_j$, and $S_k$. This set of triplets is therefore comprised of a reference sample ($S_{i}$), a sample deemed to be distant from $S_{i}$, denoted as $S_{j}$, and a sample that is similar to $S_{i}$, denoted as $S_{k}$. Note that we refer to this relationship between members of a triplet as reference sample-more distant sample-closer sample, throughout Fig. \ref{fig1} (c). Moreover, given the Random Fourier Feature representations of each sample, we compute their pairwise similarities with Euclidean distance, given as $\mathrm{D}(S_{i}, S_{j}) = \sqrt{\sum_{x=1}^{d} (S_{i}^{x} - S_{j}^{x})^2}$, where each $S_{i}^{x}$ denotes the $x$-th component of the RFF vector in sample $i$. We further use two hyperparameters ($H_s$, and $H_d$) to choose triplets that adhere to the \emph{same} and \emph{different} designations with respect to the reference sample. Hyperparameter $H_s$ controls the selection of the \emph{same} triplet member and therefore gives the cutoff point for selecting a suitable $S_{k}$ deemed to be similar enough to the reference sample, $S_{i}$ according to the computed distance $\mathrm{D}(S_{i}, S_{k})$. Alternatively, hyperparameter $H_d$ controls the threshold cutoff for a sample, $S_{j}$ to be considered different from the reference, computed according to distance $\mathrm{D}(S_{i}, S_{j}) - \mathrm{D}(S_{i}, S_{k})$. In practice, we tuned parameters $H_{s}$ and $H_{d}$ across possible quartiles of Euclidean distances between possible triplets in increments of 20. \\

\noindent \textbf{CytoCoSet Loss function.} The loss function is comprised of two terms to optimize parameters to ultimately 1) attain high classification accuracy, and 2) enforce embeddings to respect similarity patterns between members of the triplets. To promote classification accuracy, we use standard binary cross entropy, which quantifies the overlap between the true and predicted labels. In the second term, we use margin ranking loss to ensure that the triplet constraints are not violated. Further, we specify a parameter, $\alpha$ to control the trade-off between each of these two components in the loss. Putting both components together, we seek to minimize the following loss, $\ell$ in equation \ref{eq:3} as,
 
\begin{equation}
\label{eq:3}
\ell = 	\alpha \frac {1} {B} \sum_{i=1}^{B}\underbrace{ \texttt{BCELoss}(f_{\theta}(x_{i}),y_{i})}_{\text{Binary Cross Entropy}} + (1 - 	\alpha) \underbrace{\mathrm{T}(S_{i},S_{j},S_{k})}_{\text{Triplet Similarity}}.
\end{equation}

Here, $B$ denotes the batch size or the number of samples used in each batch of training, $f_{\theta}$ denotes the predicted label for sample $i$ given by the model, and $\alpha$ denotes the balance between the two terms of the loss. The learning rate and batch size are set to 0.0001 and 200 across different datasets, respectively. In the binary cross entropy term, $x_{i}$ encodes the cell $\times$ feature matrix for sample $i$, and $y_{i}$ gives the binary clinical outcome label for the $i$-th sample. Binary cross-entropy is a standard loss function component, which enables the measurement of divergence between predicted probabilities and actual binary labels, ensuring that the model's outputs are optimized to reflect accurate probabilities for each class. The triplet loss term is formulated as a margin ranking loss as $\mathrm{T}(S_{i},S_{j},S_{k})$ and is written in full as, 
\begin{equation}
\label{eq:4}
\mathrm{T}(S_{i},S_{j},S_{k})= \max \{ 0, \mathrm{dist}(S_{i},S_{k}) - \mathrm{dist}(S_{i},S_{j}) + \mathrm{h} \}
\end{equation}

 This term provides a penalty to ensure that learned embeddings for each sample do not violate known similarities between samples according to the covariates, by leveraging such rules encoded through triplets. This term is formulated as a margin ranking loss, which is well-suited for tasks requiring pairwise ranking or comparison between data instances. Intuitively, this term adds an ideal 0 to the loss if the pair of \emph{same} samples in triplet have the most similar embeddings. The parameter $\mathrm{h}$ is an offset parameter or margin to avoid trivial solutions. The $\mathrm{dist}$ used to quantify these relationships is Euclidean distance. 
\section{Results}\label{sec2}

We evaluated the performance of CytoCoSet along with several related methods on three CyTOF datasets. Our experiments explore the accuracy in predicting a clinical outcome of interest, the correlation of inferred embeddings with covariate similarities between samples, and the variation in accuracy achieved by leveraging different covariates within each dataset. 

\subsection{Datasets}
To explore the capacity of CytoCoSet to predict clinical outcome by integrating covariate information, we applied our model to three publicly available multi-sample CyTOF datasets, which we briefly introduce here. \\

\noindent \textbf{Preeclampsia dataset.} The preeclampsia dataset \cite{han2019differential} profiles 45 women throughout their pregnancies, where a subset of the women developed preeclampsia. There were 33 protein markers measured per cell. The clinical outcome to be predicted was whether the sample was from a preeclamptic or control sample. We further used gestational age (ranging between 9 and 28 weeks) as the covariate in our experiments. \\

\noindent \textbf{Preterm dataset.} The preterm CyTOF dataset \cite{peterson2021single} profiled cells extracted from cord blood in 42 women, where 17 women delivered at term and 25 women delivered prematurely. There were 36 protein markers measured per cell. Moreover, we sought to predict whether each sample was from a woman who delivered at term or at preterm. We further considered several relevant covariates across different experiments (as our method can currently only handle a single covariate). These covariates include gestational age, history of miscarriage, and preeclampsia status. \\ 

\noindent \textbf{Lung cancer dataset.} The lung cancer dataset \cite{rochigneux2022mass} contains samples from 27 patients, and measured 31 proteins per cell. We sought to predict whether or not each patient had progression-free survival (PFS). We formulated a classification problem by converting the PFS measurements to binary values based on the median value. We considered several available covariates (again, across separate experiments), including binarized age, sex, systemic immunosuppressive treatment for adverse events, and drug-related adverse events. Ultimately, we selected the administration of immunosuppressive treatment to be the default covariate to be included in analysis. \\

\noindent {\bf Details on training.} During classification experiments, we performed 30 distinct and randomized train/test splits for systematic and unbiased evaluation. To prevent against bias that could arise from different prevalence of clinical outcomes across samples in the dataset, we ensured that each training set was comprised of approximately equal proportions of samples from each clinical outcome.  

 \subsection{Related Methods Evaluated in Experiments}

In our experiments, we compared accuracy of CytoCoSet in comparison to four related methods for converting single-cell profiles into immunological summaries, including, CytoSet\cite{yi2021cytoset}, $k$-means with feature engineering\cite{stanley2020vopo}, featurization via Random Fourier Featurs\cite{rahimi2007random}, and CellCNN\cite{arvaniti2017sensitive}. We note that none of these methods take covariates into account.
\paragraph{CytoSet\cite{yi2021cytoset}} First, CytoSet is a deep learning-based method for learning representations for each sample that are predictive of clinical outcome.
\paragraph{$k$-means with feature engineering\cite{stanley2020vopo}} Under the $k$-means and feature engineering method, we cluster the cells, and then compute per-cluster frequencies, which reflect the proportion of each sample's cells assigned to each cluster. Cluster frequencies are then input into a random forest classifier to predict the clinical outcome. 
\paragraph{Random Fourier Features\cite{shan2022transparent,rahimi2007random}} Random Fourier Features (RFFs) are used to featurize each sample by projecting each cell in each sample into a higher dimensional space and then ultimately computing a per-sample representation by pooling the values of cells across each dimension (e.g. taking the max). The RFFs are then given to a Random Forest classifier to predict outcomes.  
\paragraph{CellCNN\cite{arvaniti2017sensitive}} CellCNN is a deep learning model which uses a pooling layer after the convolutional layer to combine cell-level features to generate a single representation for the entire sample. 

\subsection{Predicting Clinical Labels Across Datasets}\label{subsec1}

\begin{figure*}[!t]%
\includegraphics[width=1\textwidth]{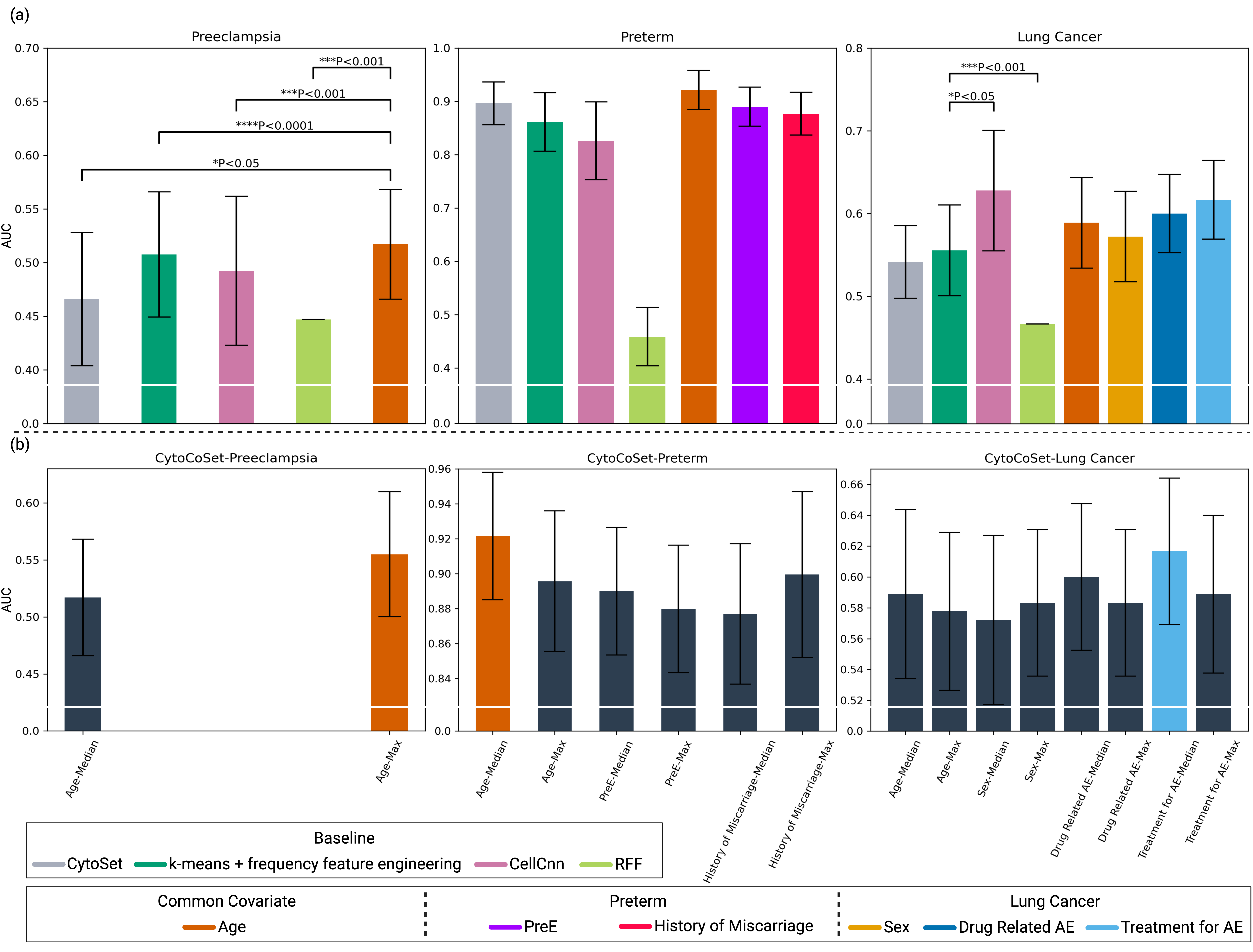}
\centering
\caption{\textbf{Classification AUC in CyTOF Datasets.} (a) CytoCoSet and baseline methods were assessed for their effectiveness in generating per-sample encodings that predict binary clinical outcomes across three CyTOF datasets (Preeclampsia, Preterm, Lung Cancer). Barplots reflect the mean AUC obtained across 30 unique train/test splits (using mean as the pooling operation to select triplets with RFFs). Error bars reflect 95\% confidence intervals around the mean. We indicated pairs of methods with statistically-significant ($p<0.05$) differences in accuracy. (b) We evaluated CytoCoSet under various choices of pooling operation in the RFF step. Results show the mean AUCs obtained by incorporating various combinations of covariate and pooling operation, denoted in labels as `covariate-pooling operation'. The covariate-pooling strategy leading the highest classification accuracy under CytoCoSet in each dataset is denoted by a \emph{non} navy blue bar.}
\label{fig3}
\end{figure*}

In our first set of experiments, we evaluated how well CytoCoSet and all related methods could predict clinical outcome. Specifically, we evaluated the hypothesis that a model which includes covariates will produce more accurate prediction of a binary clinical outcome. We computed area under the ROC curve (AUC) for classification tasks in each of the three datasets (Fig. \ref{fig3}). We further computed additional accuracy measurements, including, precision, recall, and F1 score (Supplementary Table S1).

Across datasets, the performance patterns of the various methods revealed notable insights (Fig. \ref{fig3} and ROC curves shown in Supplementary Fig S1). For example, in the lung cancer dataset, CytoSet achieved an AUC of 0.54, and CytoCoSet was able to further improve on this accuracy using the systemic immunosuppressive treatment for adverse events as the covariate, achieving an AUC of 0.62. Alternatively, when using drug-related adverse or age as covariates, CytoCoSet achieved stronger accuracy than CytoSet. While the CellCNN approach technically achieved the highest AUC, it also showed substantial variance in performance across trials, suggesting a lack of robustness of such an approach. In the preeclampsia dataset, CytoCoSet achieved an AUC of 0.55 using binarized age as the covariate and max pooling in the RFF process (Fig. \ref{fig3} (b)), which surpassed CytoSet's AUC of 0.47. In particular, each baseline method had statistically significantly worse performance across trials than that achieved by CytoCoSet. In the preterm dataset, by construction, gestational age is highly correlated with the term of preterm status, so we anticipated the covariate would help performance. While CytoSet achieved an AUC of 0.9, CytoCoSet attained an AUC of 0.92. Conversely, the featurization via RFFs exhibited the lowest AUC of 0.46, followed by frequency feature engineering and CellCNN, which achieved AUCs of 0.86 and 0.83, respectively. Additional accuracy metrics were computed and are shown in Supplementary Tables S1 and S2.

Across multiple datasets, various covariates exhibited varying degrees of effectiveness in enhancing prediction accuracy. For example, age is a common covariate across datasets, but it is only helpful for clinical outcome prediction in some datasets. Notably, in the preeclampsia dataset, including age information yields a substantial increase in AUC. Conversely, in the preterm dataset, the impact of gestational age as a covariate is comparatively modest, and only marginally increases AUC. The noteworthy enhancement of prediction accuracy with age as a covariate as observed in the preeclampsia dataset suggests that immune profiles augmented with age could produce a more accurate predictor of preeclampsia. 

\subsection{Evaluating Alignment of Embedding Vectors and Covariates}\label{subsec2}
\begin{figure}[!t]%
\includegraphics[width=0.8\textwidth]{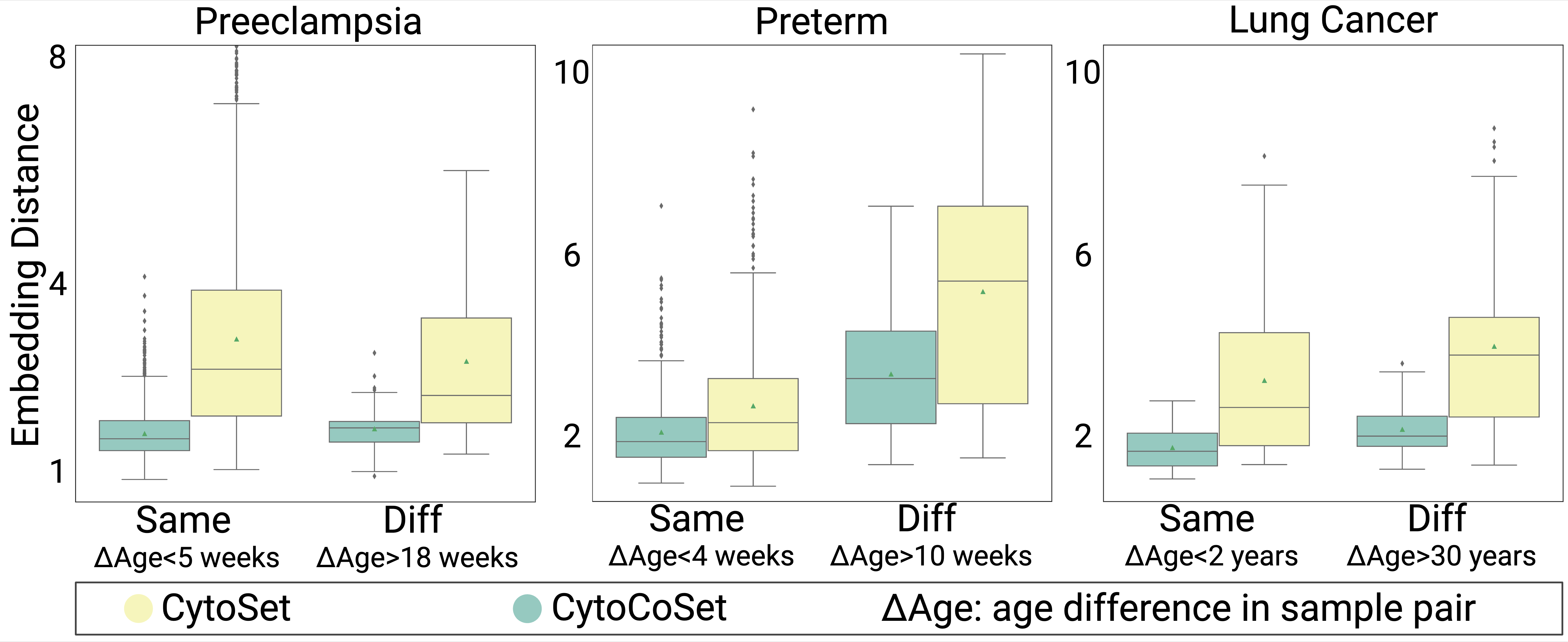}
\centering
\caption{\textbf{Quantifying Alignment of Embedding Vectors with Covariates.} Boxplots visualize distances computed between pairs of samples with the same age (denoted as `Same') and between those with different-age (denoted as `Diff') under CytoSet (yellow) and CytoCoSet (green) approaches. The green triangle in each boxplot represents the mean embedding distance.}
\label{fig4}
\end{figure}

We designed an experiment to evaluate how well the learned embedding vectors align with the age covariate that is available in each dataset. We expected that CytoCoSet should produce embeddings that are most similar between samples from donors of the same age. To test this, we extracted embedding vectors for each sample based on the second-to-last layer of the network. For many random pairs of samples (specifically 66, 190, and 120, in the lung cancer, preeclampsia, and preterm datasets, respectively), we computed the Euclidean distance between the respective per-sample embeddings returned in the second-to-last layer. This experiment was repeated over thirty trials to generate a distribution of embeddings distances between pairs of samples that both had the same (`Same') and different (`Diff') ages, under both CytoCoSet and CytoSet. We visualized these distributions in each dataset in Fig. \ref{fig4}. We observed that the distances between the embeddings for samples from donors with the same age computed under CytoCoSet tended to be lower than the covariate-agnostic variant, CytoSet, across the three CyTOF datasets. Note that we specified an age difference in each dataset required for two samples to be considered as having different ages (denoted as`$\Delta \text{Age}$'). Analyses of pairwise embedding vector distances in the \emph{same} and \emph{different} comparisons revealed consistent trends across the three datasets. Notably in the preterm and lung cancer datasets, embeddings obtained under CytoCoSet between pairs of samples with the same age tend to be much more similar than embeddings obtained between samples of different ages. In the preeclampsia dataset, CytoSet erroneously showed smaller embedding vector differences between samples from donors of different ages, while CytoCoSet exhibited the correct behavior.  
% In the preterm dataset, same-age was defined as pairs with less than four weeks difference and different-age as greater than ten weeks. Conversely, within the lung cancer dataset, same-age was less than two years difference and different-age was more than thirty years. In the preeclampsia dataset, same-age was less than five weeks difference and different-age was more than eighteen weeks.

\subsection{Sensitivity of Parameters In the Loss Function}\label{subsec3}
\begin{figure*}[!t]%
\includegraphics[width=1\textwidth]{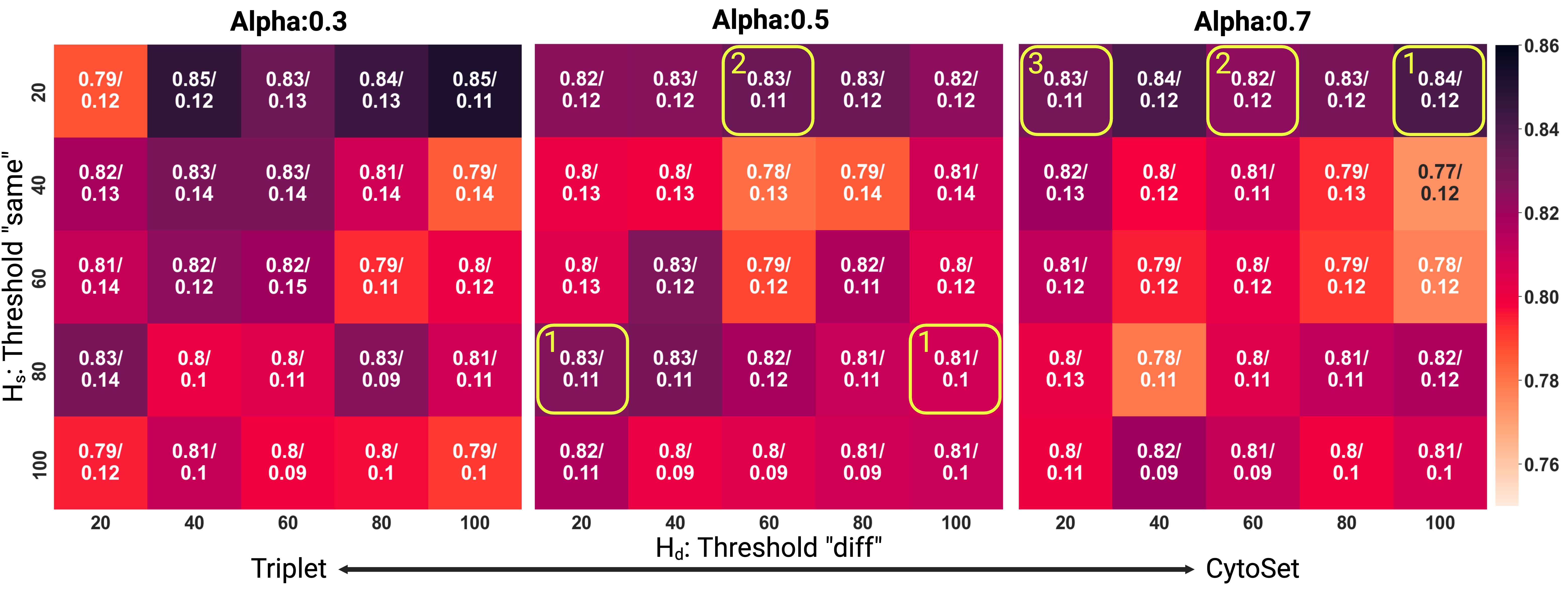}
\centering
\caption{\textbf{Sensitivity Analysis of Parameters within the Loss Function.} We systematically varied the three model hyperparameters, $\alpha$, `same threshold' ($H_{s}$) and `diff threshold` ($H_{d}$) and visualized the mean AUC/standard deviation over ten trials with different train/test splits. Ten trials were run with different train/test splits, and yellow squares denote the optimal hyperparameter combinations across the ten trials. Each heatmap grid also denotes the number of trials that a particular parameter combination was optimal in.}\label{fig5}
\end{figure*}

The tuneable parameters in the model include the `same threshold', $H_{s}$ `diff threshold`, $H_{d}$ and $\alpha$. In practice, lower quartile values for $H_{s}$ and $H_{d}$ involve more stringent triplets with closer $(S_{i}, S_{k})$ distances and larger $(S_{i}, S_{j})$ distances. Next, $\alpha$, controls the relative contributions of the binary cross entropy term and the triplet term in the loss function. We systematically varied all three parameters and visualized the resulting mean/standard deviation of the test-set AUC, obtained across ten trials in the preterm dataset (Fig. \ref{fig5}). All experiments here constructed triplets through median pooling based on Random Fourier Features (RFFs). This experiment revealed that the same threshold, $H_{s}$ is the more sensitive parameter. That is, choosing triplets more appropriately for the same threshold will have a more productive effect on the loss. We varied parameters $H_{s}$ and $H_{d}$ along with $\alpha={0.3, 0.5, 0.7}$ across ten trials. The best results were obtained for $\alpha=0.5$ and $\alpha=0.7$. These optimal hyperparameter combinations were denoted in  Fig. \ref{fig5} with yellow squares and numbers to show the number of the ten trials for which the particular parameter combination was optimal. Overall, CytoCoSet revealed some consistency in optimal parameter values across trials, but there was still some variation that could be investigated in future studies. 

\subsection{Examining Effects of Different Covariates}\label{subsec4}
\begin{figure}[!t]%
\includegraphics[width=0.8\textwidth]{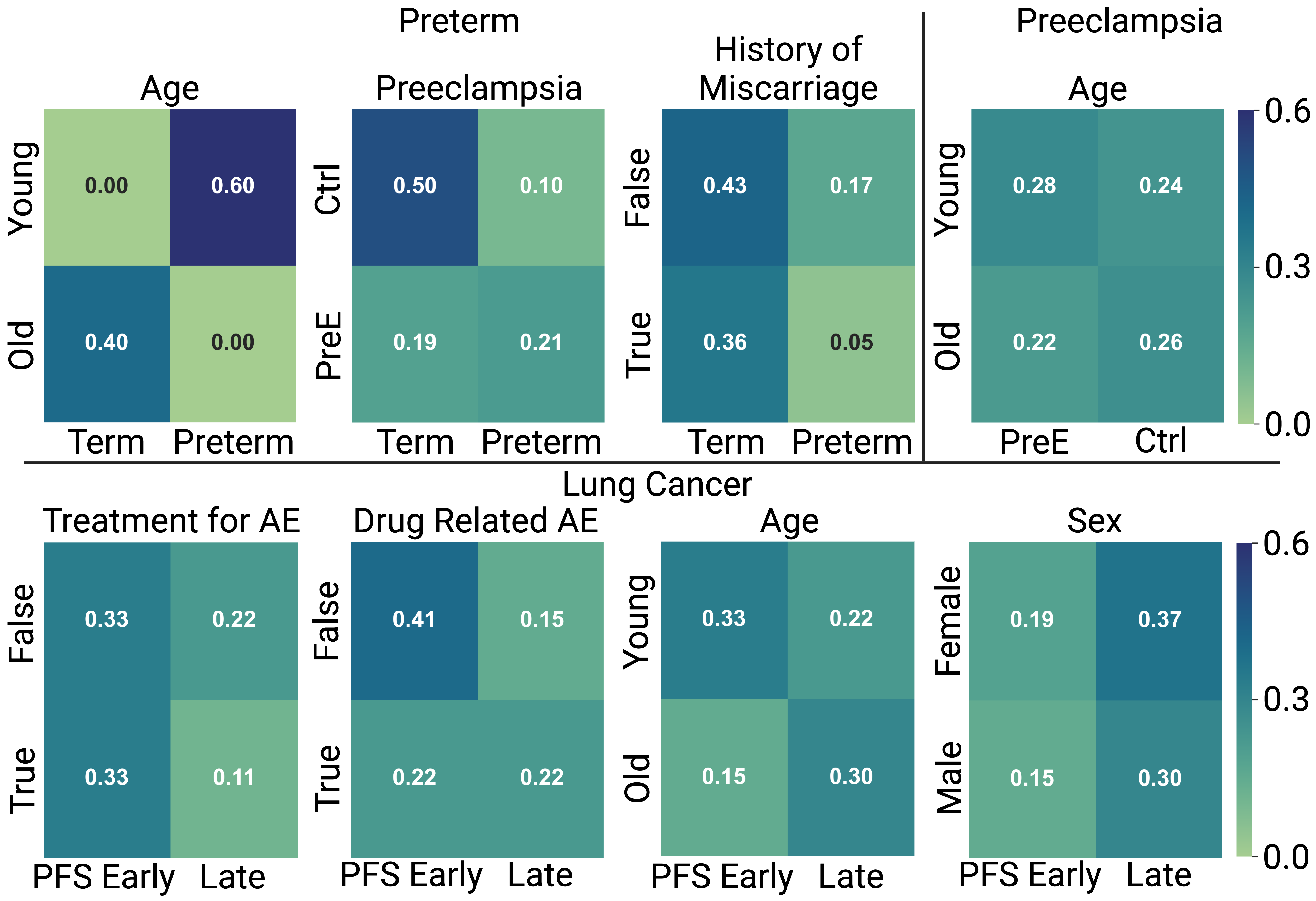}
\centering
\caption{\textbf{Examining correlations between the covariate and the outcome to be predicted.} The heatmap shows the correlation (light to dark indicating low to high correlation) between the label to be predicted and several possible covariates in each dataset. For example, in the preterm dataset there is an expected high correlation between the preterm outcome and (gestational) age, which is expected.}\label{fig6}
\end{figure}

Each dataset typically includes multiple covariates per patient. We anticipate that every covariate will not exhibit the same degree of correlation with the outcome. Our assumption is that embedding vectors should be more similar for samples with similar covariates is only valid if such covariates are aligned with the label used to optimize the network parameters. Here, we explored how different covariates affected classification accuracy in the preterm and lung cancer datasets. The heatmaps in  Fig. \ref{fig6} show the correlation between the label to be predicted and each of the available covariates. In the preterm dataset, we observed a strong correlation between (gestational) age and the preterm condition, indicating that this is a good candidate covariate to use in learning embeddings. Alternatively, the history of miscarriage covariate correlated poorly with the preterm label. Alternatively in the lung cancer dataset, none of the measured covariates,  including, treatment for AE, drug related AE, age, or sex had significantly high correlations with progression-free survival, except for Drug Related AE, which exhibited a moderate correlation. 

\begin{figure}[!t]%
\includegraphics[width=0.8\textwidth]{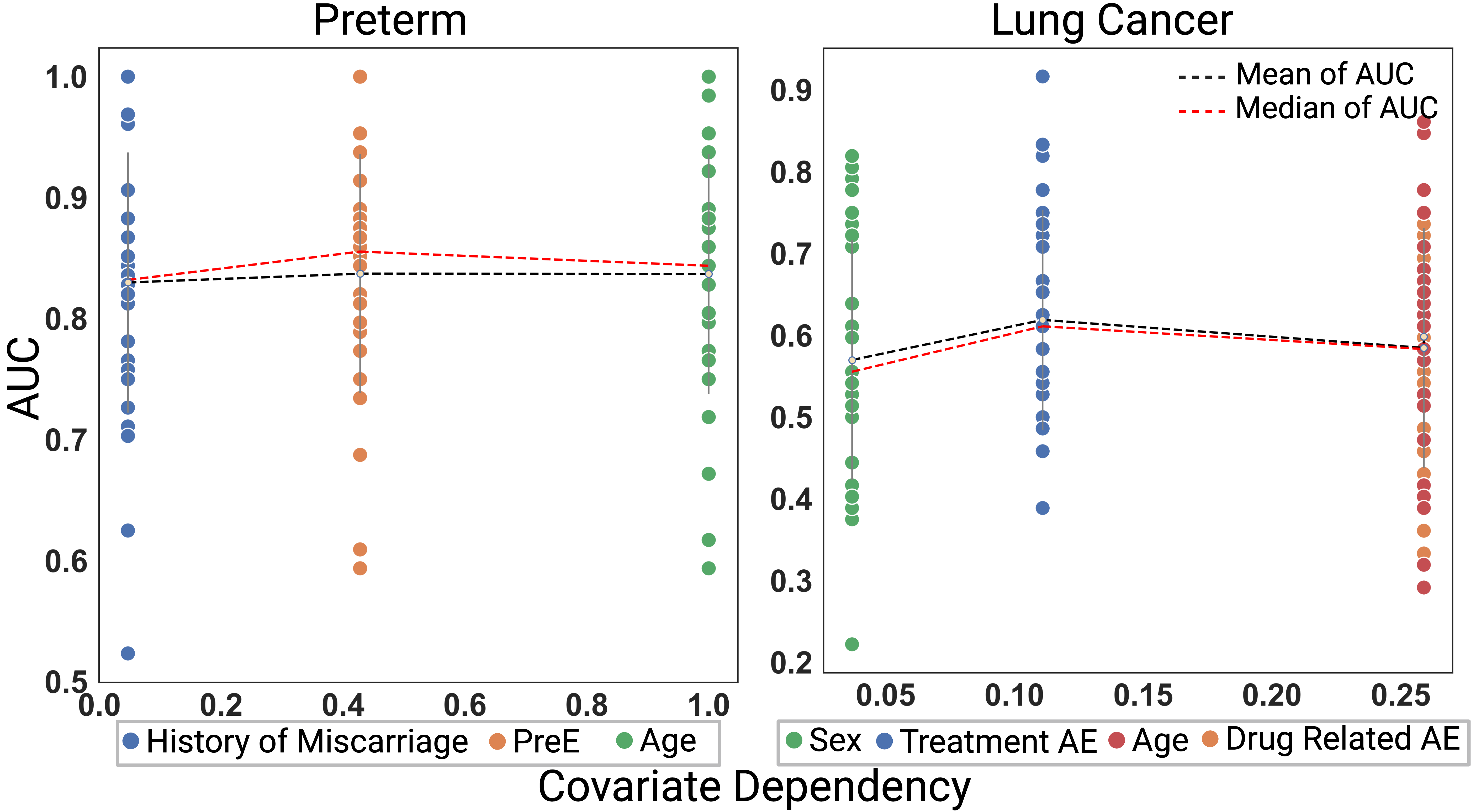}
\centering
\caption{\textbf{Evaluation of how covariate dependency impact accuracy in the preterm and lung cancer datasets.} We examined how the extent to which each of the possible covariates and the outcome to be predicted influence outcome prediction in the preterm (PreE is used to denote the covariate of preeclampsia status) and lung cancer datasets (AE denotes adverse events). Higher values of `covariate dependency' (horizontal axis) imply stronger alignment between a covariate and the outcome. CytoCoSet was run over 30 trials, under each covariate. Here, covariates are ordered as a function of increasing covariate dependency score. We report AUC (dots, vertical axis) as the metric of success to quantify how useful the embeddings were in the classification task. }\label{fig7}
\end{figure}

To explore these patterns more rigorously, we designed a statistic to efficiently identify which covariates and outcomes are most related and could be leveraged for improved model training. Given binary values for a particular covariate and a binary outcome, we can construct a two-by-two square matrix representing the correlation between each binary outcome and each of the binary values of the covariate. We then define notation to denote the top left, top right, bottom left, and bottom right quadrants are denoted as ${D}_0$, ${D}_1$, ${D}_2$, and ${D}_3$, respectively. Given this constructed matrix, we can compute the statistic to quantify alignment between a covariate and an outcome as,  
\begin{equation}
\label{eq:5}
\text{Covariate Dependency} = \lvert( {D}_1 + {D}_2 ) - ( {D}_0 + {D}_3 ) \lvert
% abs(abs(2+1)-abs(0+3))
\end{equation}
Covariate Dependency is a statistic that can be computed to capture the difference between two sums of the diagonal, which correspond to combinations of a given covariate level and an outcome. In an ideal setting, this difference should be large indicating strong alignment of an outcome with a particular covariate. We set up prediction tasks in each dataset under each possible covariate and plotted the distribution of AUCs (each trial represented by a dot) over 30 trials in Fig. \ref{fig7} as a function of covariate dependency. We expect an increase in AUC as a function of increasing covariate dependency. Thus, a higher AUC is anticipated as the difference becomes more pronounced, indicating enhanced discriminative ability and predictive accuracy in the model. The results suggest that mean or median AUC (shown by the dotted line) is generally larger for more informative covariates, or those with larger covariate dependency. Specifically, in the preterm experiment, the covariate (gestational) 'age' exhibits the highest degree of significance, as evidenced by its high covariate dependency value, consistent with the findings illustrated in Fig. \ref{fig3}. This observation aligns with the notion that prematurity, defined as being born with a gestational age of less than 37 weeks, has a significant effect on the neonate immune system \cite{peterson2021single}. In the lung cancer dataset the 'drug-related adverse events' covariate has the highest covariate dependency, which also has a relatively high AUC in Fig. \ref{fig3}. Several studies have emphasized how drug-related adverse events experienced in patients receiving immunotherapy ultimately impact treatment outcomes in lung cancer\cite{morimoto2021immune, shankar2020multisystem, rochigneux2022mass, wang2016role, zhu2021immune}.

\subsection{CytoCoSet Captures Between-Sample Variation}\label{subsec5}
\begin{figure}[!t]%
\includegraphics[width=0.8\textwidth]{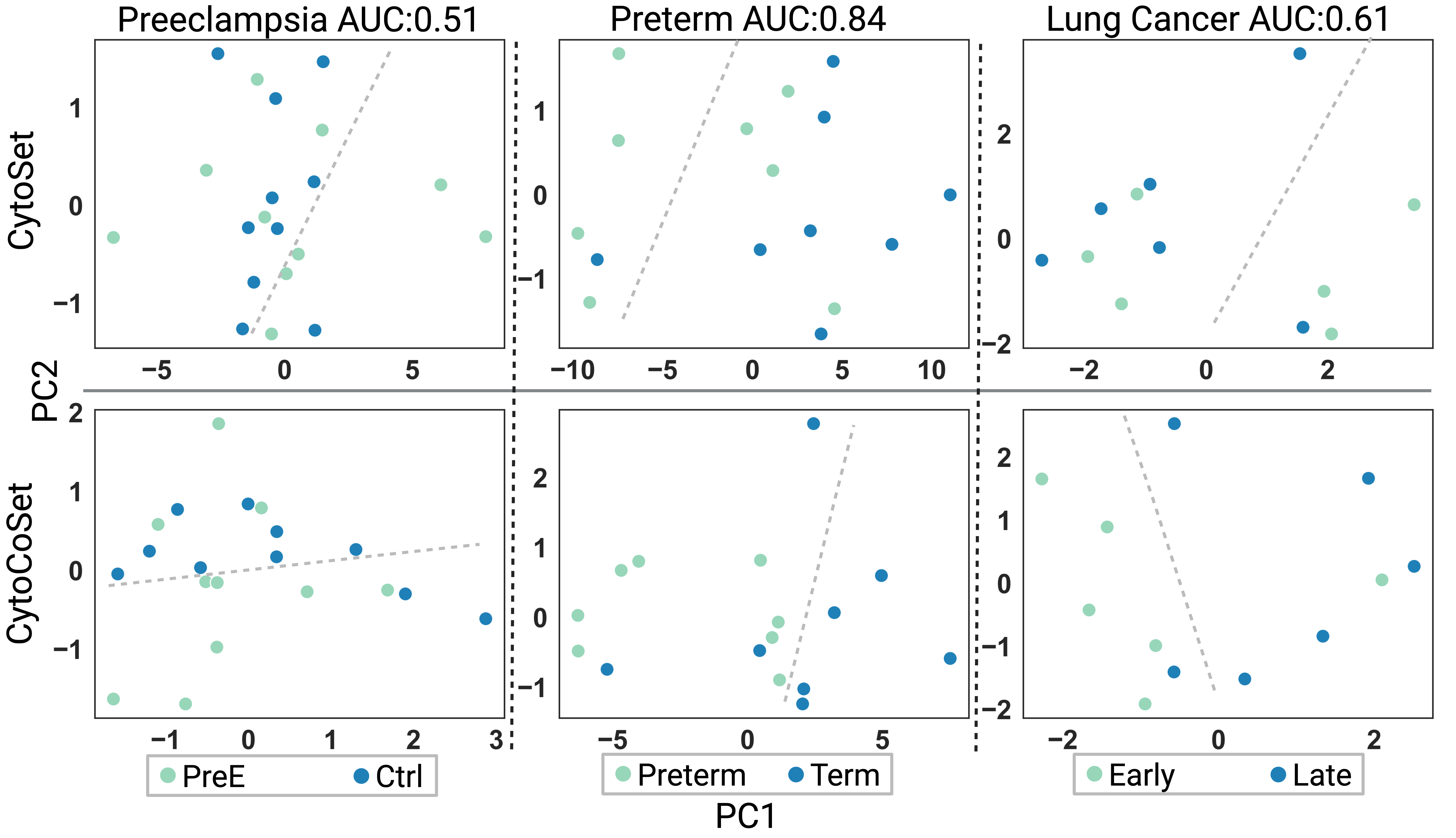}
\centering
\caption{\textbf{Visualizing Samples in 2d Based on Learned Embeddings.} PCA was used to project samples in two-dimensions with PCA, based on the learned embedding coordinates under CytoSet (top) and CytoCoSet (bottom). Thirty classification trials were run and we visualized projections of test set samples corresponding to the trial achieving the median AUC (denoted AUCs are those obtained with CytoCoSet). Dots represent samples and are colored by the class label. Gray dotted lines represent a possible decision boundary that could be used to separate the samples correctly.}\label{fig8}
\end{figure}

We sought to understand the extent to which the learned embeddings captured variation between samples. That is, we expected that information-rich embeddings could be used as input to a classifier or dimension reduction technique to reveal separation according to outcome. We ran 30 classification trials with different train/test splits to predict outcomes based on embedding vectors learned with Cytoset (top) or CytoCoSet (bottom). Fig. \ref{fig8} shows two-dimensional projections of each samples based on their learned embeddings under the trial that obtained the median AUC across the 30 trials. The gray dotted line shows a possible decision boundary that could separate points in different classes, as achieved using the Nu-Support Vector Classification. Overall, the results suggest that CytoCoSet is superior in learning embeddings that correspond accurately separate samples according to clinical outcome.

\section{Discussion}

The paper defines a robust method, CytoCoSet, for incorporating patient-specific clinical covariates to learn per-sample featurizations to summarize each patient's cellular landscape through a learned featurization or embedding vector. As a first key contribution, CytoCoSet proposes a set-based encoding method to robustly take into account per-sample outcome labels to be predicted with additional patient covariates. CytoCoSet expands existing deep learning set-based encoding methods, such as CytoSet  \cite{yi2021cytoset}, to learn per-sample featurizations accounting for both donor-level covariates and immune signatures for clinical prediction tasks. As an algorithmic contribution, we enhanced the conventional set-based encoding formulation of the loss function proposed for learning model parameters in CytoSet by introducing a triplet penalty term. This term penalizes learned embedding vectors from being substantially different between samples with similar covariates. Finally in practice, the tunable parameter, $\alpha$ enables flexible control over the contributions of the binary cross-entropy and triplet terms in the loss function, enhancing the overall effectiveness of the training process and user experience. 

CytoCoSet has proven that the augmentation of immune features and donor level covariates has the capacity to greatly improve the prediction of clinical outcome. Such an approach infers complex patterns that may exist between immune signatures, the outcome to be predicted, and other measured covariates.  CytoCoSet and future related methods can successfully infer per-sample embeddings or featurizations based on such information to reveal novel patient subgroups that could inspire more specialized treatment strategies. For example, Refs. \cite{phongpreecha2024single,he2024phenotype} have introduced cutting-edge single-cell immune profiling data with additional information about behaviors and co-morbidities that could be jointly integrated to better diagnose severity and progression rates of neurodegenerative diseases. Such datasets, in addition to those introduced in the paper reveal key settings where an algorithm like CytoCoSet could be immensely beneficial.    

However, there are still substantial opportunities to expand this methodology to accommodate the additional and ubiquitous clinical setting where there are multiple covariates measured per donor that could be leveraged when computing featurizations. A future direction is therefore to extend the CytoCoSet model to simultaneously incorporate multiple patient-level covariates by implementing alternative formulations of the triplet term or new penalty terms to enforce certain relationships between samples. For example, a possible approach is to modify the training process so that the model infers and leverages information about rank-ordered correlations between clinical outcomes and covariates. Specifically, these correlations can be used to create a new composite penalty term that weights the contribution of covariates in the loss based on their correlation with the clinical outcome. Additionally, future work should address methodology to impute missing covariates and ensure that the algorithm's scalability enables it to be applied to even larger datasets with potentially millions of cells (scalability explored in Supplementary Fig. S2). Finally, biological interpretability is an additional aspect that should be improved in future implementations of CytoCoSet. Specifically, future work can leverage advancements in explainability \cite{lundberg2017unified} to communicate the key cells driving the prediction. 

Overall, CytoCoSet pioneers a new direction in formulating a novel deep learning approach to augment complex immune signatures gleaned through single-cell profiling with clinical covariates. This integration of information enables more robust prediction of clinical phenotype and the potential to improve diagnostics and treatment strategies.

\section{Declarations}

\noindent \textbf{Ethics approval and consent to participate.} Not applicable.\\

\noindent \textbf{Consent for publication.} Not applicable.\\

\noindent \textbf{Availability of data and material.} The three datasets are accessible, organized, and published in Zenodo under the titles Preeclampsia (DOI: \href{https://zenodo.org/records/10659650}{10.5281/zenodo. 10659650}), Preterm (DOI: \href{https://doi.org/10.5281/zenodo.10660080}{10.5281/zenodo.10660080}), and Lung Cancer (DOI: \href{https://doi.org/10.5281/zenodo.10659930}{10.5281/zenodo.10659930}). The CytoCoSet software is available as open-source code on GitHub at the following URL: \url{https://github.com/ChenCookie/cytocoset}. Other detail on the
evaluation is provided in the \href{https://github.com/ChenCookie/cytocoset/blob/main/Supplementary_Material/CytoCoSet_Supplementary.pdf}{supplementary material}.\\

\noindent \textbf{Competing interests.} Not applicable.\\

\noindent \textbf{Funding.} This work is supported by The National Institutes of Allergy and Infectious Diseases of the National Institutes of Health under the award number 1R21AI171745-01A1 (NS).\\

\noindent \textbf{Author contributions.} CJ.C and N.S conceptualized the project. CJ.C implemented the methodology, acquired single-cell datasets, and ran all experiments. H.Y assisted with data analysis and implementation. N.S and CJ.C wrote the manuscript with input from all authors. N.S acquired funding and supervised the project.\\

\noindent\textbf{Acknowledgements.} Not applicable.\\

\bibliographystyle{ieeetr} %alpha, apalike, ieeetr
\bibliography{bibliography.bib}

\end{document}